\documentclass[letterpaper]{article} 
\usepackage{aaai25}  
\usepackage{times}  
\usepackage{helvet}  
\usepackage{courier}  
\usepackage[hyphens]{url}  
\usepackage{graphicx} 
\usepackage{natbib}  
\usepackage{caption} 
\usepackage{algorithm}
\usepackage{algorithmic}
\usepackage{amsmath}
\usepackage{booktabs}
\usepackage{tabularx}
\usepackage{multirow}
\usepackage{amssymb}
\usepackage{siunitx}
\usepackage{rotating}
\usepackage{xcolor}
\usepackage{enumitem}
\usepackage{subcaption}

\usepackage{newfloat}
\usepackage{listings}
\DeclareCaptionStyle{ruled}{labelfont=normalfont,labelsep=colon,strut=off} 
\floatstyle{ruled}
\newfloat{listing}{tb}{lst}{}
\floatname{listing}{Listing}
\title{Deep Generative Model for Mechanical System Configuration Design}
\author {
    Yasaman Etesam\textsuperscript{\rm 1,\rm 2},
    Hyunmin Cheong\textsuperscript{\rm 1},
    Mohammadmehdi Ataei\textsuperscript{\rm 1},
    Pradeep Kumar Jayaraman\textsuperscript{\rm 1}
}
\affiliations {
    \textsuperscript{\rm 1}Autodesk Research,
Toronto, Ontario, Canada\\
    \textsuperscript{\rm 2}School of Computing Science, Simon Fraser University, Burnaby, BC, Canada\\
    yetesam@sfu.ca,
    \{hyunmin.cheong, mehdi.ataei, pradeep.kumar.jayaraman\}@autodesk.com
}

\usepackage{bibentry}

\begin{document}

\maketitle

\begin{abstract}
Generative AI has made remarkable progress in addressing various design challenges. One prominent area where generative AI could bring significant value is in engineering design. In particular, selecting an optimal set of components and their interfaces to create a mechanical system that meets design requirements is one of the most challenging and time-consuming tasks for engineers. This configuration design task is inherently challenging due to its categorical nature, multiple design requirements a solution must satisfy, and the reliance on physics simulations for evaluating potential solutions. These characteristics entail solving a combinatorial optimization problem with multiple constraints involving black-box functions. To address this challenge, we propose a deep generative model to predict the optimal combination of components and interfaces for a given design problem. To demonstrate our approach, we solve a gear train synthesis problem by first creating a synthetic dataset using a domain-specific language, a parts catalogue, and a physics simulator. We then train a Transformer-based model using this dataset, named \emph{GearFormer}, which can not only generate quality solutions on its own, but also augment traditional search methods such as an evolutionary algorithm and Monte Carlo tree search. We show that GearFormer outperforms such search methods on their own in terms of satisfying the specified design requirements with orders of magnitude faster generation time. Additionally, we showcase the benefit of hybrid methods that leverage both GearFormer and search methods, which further improve the quality of the solutions. 
\end{abstract}

%

\section{Introduction}

Configuration design of mechanical systems in engineering \cite{mittal1989towards, wielinga1997configuration} is a time-consuming task that relies on the domain expertise of engineers. Typically, engineers manually select the optimal combination of components to meet multiple design requirements based on their experience and knowledge, often leading to sub-optimal solutions that negatively impacts product development \cite{suh1990, dym1994engineering}.

Computationally, configuration design can be defined as a combinatorial optimization problem with categorical design variables (e.g., component choices) and design requirements expressed as objectives and constraints \cite{levin2009combinatorial}. Evaluating these objectives or constraints frequently necessitates the use of black-box physics simulation tools. Due to these aspects, configuration design problems have traditionally been approached using gradient-free, meta-heuristic optimization methods like evolutionary algorithms \cite{deb2003multi, angelov2003automatic, grignon2004ga, cheong2019configuration} or simulated annealing \cite{schmidt1998optimal, shea1997shape}. In certain problems, search methods such as Monte Carlo tree search \cite{zhao2020robogrammar, luo2022alphatruss} or heuristic search \cite{campbell1998agent, piacentini2020multi, zhao2020robogrammar} have been employed. While these methods have shown some success, they are computationally expensive and time-consuming, limiting their practical adoption by engineers who ideally seek real-time recommendation of solutions to perform multiple design iterations within constrained time-frames \cite{cagan2002creating}.

The need for a fast, interactive tool motivated our investigation into deep generative models for mechanical system configuration design. Recent progress in generative models, particularly Transformers~\cite{vaswani2017attention}, offers a promising approach to the problem. Originally designed for natural language processing, Transformers excel at generating complex sequences and have proven effective in solving combinatorial problems, such as AlphaFold's protein structure predictions \cite{varadi2022alphafold, jumper2021highly}. In contrast to traditional methods, these generative models can instantly predict solutions given the user input.

Mechanical systems have \emph{domain-specific languages} similar to the natural language, consisting of grammar and lexicon. Just as sentences must follow syntactic rules with comprehensible words, valid mechanical configurations must consist of specific parts that adhere to compatibility constraints. For example, constructing a basic mechanical linkage involves a sequence such as [pivot point] - [connecting rod] - [pin joint]. Each component must be compatible with the next for the linkage to be physically realizable and function correctly. While this is a simple example, more complex systems follow similarly structured rules. This parallel further motivated us to investigate applying Transformer models for generating valid and functional mechanical designs.

\begin{figure*}
    \centering
    \includegraphics[width=1\linewidth]{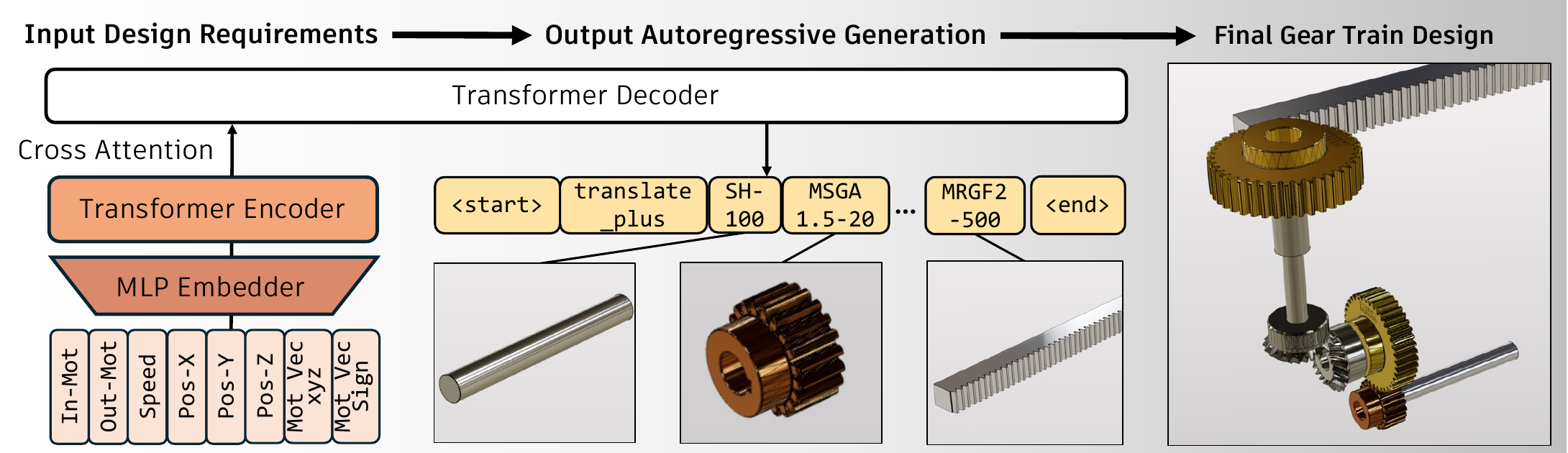}
    \caption{GearFormer is based on the Transformer architecture. It includes an encoder module that processes the input design requirements that have been embedded by a multi-layer perceptron (MLP). These input embeddings are consumed by the Transformer decoder module via cross-attention to generate the gear train sequence that satisfies the requirements.}
    \label{fig:model_arch}
\end{figure*}

\begin{figure}
    \centering    \includegraphics[width=0.8\linewidth]{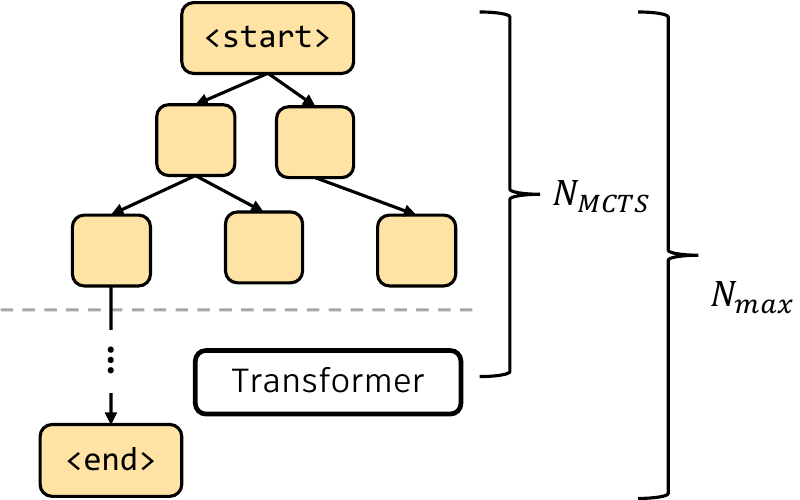}
    \caption{A hybrid method that combines MCTS to explore the first few critical tokens in a sequence and a Transformer-based model to complete the sequence for evaluation.}
    \label{fig:monte_carlo_search}
\end{figure}

To explore this premise, a novel Transformer-based generative model is developed for gear train synthesis. Gear trains are complex and ubiquitous mechanical systems, consisting of various types of components and interfaces that transmit motion while adjusting speed or torque. Designing gear trains that satisfy multiple requirements demands significant domain expertise and time, thus serves as an ideal problem for tackling the challenges of configuration design. To train such a Transformer-based model, we developed a method to synthetically generate datasets for mechanical configuration design problems because such datasets are not publicly available. Our main contributions are as follows:

\begin{itemize}
    \item The introduction of a Transformer-based model named \emph{GearFormer} (Figure \ref{fig:model_arch}), to solve an archetypal mechanical configuration design problem in gear train synthesis. 
    \item Augmenting traditional search methods such as Estimation of Distribution Algorithm (EDA) and Monte Carlo tree search (MCTS) with GearFormer, e.g., Figure \ref{fig:monte_carlo_search}, to obtain higher quality solutions than their own.
    \item The creation of the GearFormer dataset, the first dataset for gear train synthesis, created with a domain-specific language and augmented with physics simulations.
    \item The development of a physics simulator capable of evaluating multiple requirements of a gear train design.
\end{itemize}

\begin{figure*}[!h]
    \centering
    \includegraphics[width=1\linewidth]{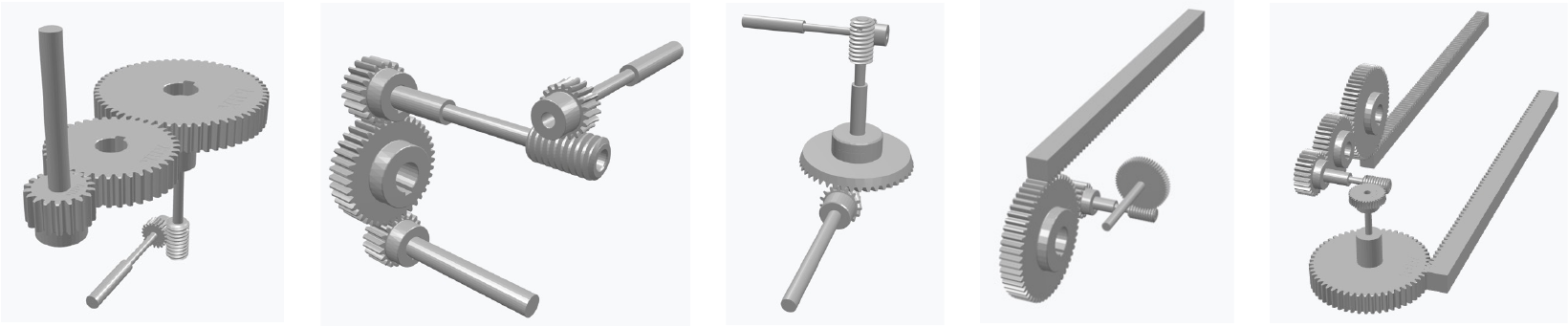}
    \caption{Gear train designs generated with GearFormer}
    \label{fig:designs}
\end{figure*}

We show that GearFormer efficiently generates high-quality designs (e.g., Figure \ref{fig:designs}) that meet design requirements compared to traditional search methods. Moreover, we demonstrate that it can augment search methods to produce better solutions in a shorter amount of time than their own. A link to our code, data, and supplementary material referenced in this paper can be found at \url{https://gearformer.github.io/}. The website also features video demos that illustrate how GearFormer can be used in design workflows.

\section{Related Work}

Configuration design has been a subject of extensive research in engineering design automation. Various computational approaches have been developed to find optimal component combinations that satisfy design requirements, including evolutionary methods. \citet{deb2003multi} proposed a multi-objective evolutionary algorithm for solving complex engineering design problems. \citet{angelov2003automatic} developed a fuzzy rule-based evolutionary approach for robot configuration design. \citet{grignon2004ga} applied genetic algorithms to optimize product family designs. \citet{cheong2019configuration} used EDAs for configuration design of suspension systems. 

Tree search methods have also shown promise in solving configuration design problems. \citet{campbell1998agent} introduced an agent-based approach using heuristics for conceptual design. \citet{piacentini2020multi} applied heuristic search to a multi-speed gearbox design problem. \citet{zhao2020robogrammar} developed a graph grammar-based approach with MCTS for robot design. \citet{luo2022alphatruss} employed MCTS for optimizing truss structures.

Incorporating design knowledge in the form of grammars or heuristics has been a key strategy to improve the efficiency and effectiveness of configuration design methods. \citet{schmidt1998optimal} used simulated annealing with a grammar-based representation for mechanical configuration design. \citet{shea1997shape} applied shape grammars to generate structural designs. \citet{piacentini2020multi} and \citet{zhao2020robogrammar} also incorporated design knowledge through heuristics and grammars in their approaches.

Recent advances in generative models have primarily focused on design ideation, which typically precedes configuration design. \citet{regenwetter2022deep} reviewed deep generative models in engineering, while \citet{chen2024towards} and \citet{ataei2024elicitron} explored large language models and ontologies for conceptual generation and requirements elicitation. \citet{chen2024designfusion} integrated generative models for early-stage design. Research into AI-augmented design \citep{thoring2023augmented} and sketch-based inputs \citep{zhang2023generative} with tools like DesignAID \citep{cai2023designaid} advance the design ideation process but are still limited in practical configuration design.

\section{Problem}

\subsection{Configuration Design of Mechanical Systems}

We address a class of configuration design problems where the design solution can be represented as a sequence. For example, designing an open-loop kinematic mechanism with single input and single output motions/forces would fall under this category. A design solution can be expressed as: 
\begin{equation}
X = (\mathcal{S}, C_1, I_{1,2}, C_2, ..., C_i, I_{i,i+1}, C_{i+1}, ..., C_N, \Sigma)
\label{eq:1}
\end{equation}
where $\mathcal{S}$ is a start symbol, $C_i$ are components, $I_{i, i+1}$ are interfaces between the preceding and the following components, $\Sigma$ is a terminal symbol, and $N$ is the number of components in the sequence. The possible values of $C_i$ are typically based on the list of off-the-shelf components available to the engineer and the possible values of $I_{i, i+1}$ are based on the allowed interfacing between $C_i$ and $C_{i+1}$. $X$ must be chosen such that it conforms to a set of rules, $\mathcal{R}$. For example, if $C_1$ was a rack component, $C_2$ must be a spur gear component that can mesh with $C_1$ and $I_{1, 2}$ must convey the corresponding meshing. The possible values and set of rules for a particular application domain constitute the lexicon and grammar of a \emph{domain-specific language} (DSL). 

Solving a configuration design problem involves finding an optimal sequence $X$ that minimizes a particular objective function while satisfying multiple design requirements posed as constraints. In general, evaluating the objective function or constraints involves a black-box solver such as a physics simulator. Due to the combinatorial nature of the problem and non-differential functions, solving the problem with traditional computational methods is challenging. 

\subsection{Gear Train Synthesis Problem}
To investigate solving a configuration design problem with a generative model approach, we focus on the gear train synthesis problem. A gear train with single input/output motions can be represented as a sequence defined in Eq. \eqref{eq:1}, an enumerated collection of gear components and their interfaces.

Design requirements considered for gear trains include the cost, weight, output speed/torque ratio, output motion position/vector, and input/output motion type conversion. We used the total weight of a gear train as our objective function, which also serves as a good proxy for the cost. Also, since a torque ratio is simply an inverse of a speed ratio for a gear train ignoring its efficiency, it was omitted for this problem. The rest of the requirements were posed as constraints. We also ensure that a gear train is physically realizable by conforming to the domain-specific grammar and having no interference among components. The optimization problem for gear train synthesis can be formulated as follows.
\begin{equation}
\begin{alignedat}{4}
& \underset{X}{\text{min}}\quad&& f_w = \sum_{i=1}^{N}(w_{C_i}), && C_i \in X  \\ 
& \text{s.t.} && \Tilde{s} - s(X) = 0;  && |\widetilde{\vec{p}} - \vec{p}(X)| = 0 \\
& && \widetilde{\vec{m}} \cdot \vec{m}(X) - 1 = 0 && \\
& && \Tilde{\tau}_{\text{in}} - \tau_{\text{in}}(X) = 0; && \Tilde{\tau}_{\text{out}} - \tau_{\text{out}}(X) = 0 \\
& && g_{\text{grammar}}(X) = 0; && g_{\text{interfere}}(X) = 0 \\
\end{alignedat}
\label{eq:gear_opt}
\end{equation}

$f_w$ is the total weight of the gear train, $w_{C_i}$ is the weight of each component $C_i$ that is a member of the gear train sequence $X$, $s$ is the speed ratio, $\vec{p}$ is the output position in $x, y, z$ coordinates , $\vec{m}$ is the output motion vector where the nonzero index indicates the coordinate and its sign indicates the direction, $\tau_{in}$ is the input motion type, and $\tau_{out}$ is the output motion type. Those denoted with \emph{tilde} are targets provided by the engineer. Operators $g_{\text{grammar}}$ and $g_{\text{interfere}}$ return $0$ if $X$ conforms to the grammar and does not have interference among parts, respectively. The rest of operators that are functions of $X$ are evaluated using a physics-solver. 

\section{Proposed Approach}

\subsection{Transformer-based Model}
To solve the configuration design problem, we propose a generative model that takes the representation of the design requirement constraints as an input encoding vector $\mathcal{E}$ and outputs a sequence $X$ conditioned based on that input while also minimizing the objective function $f(X)$. In other words, we aim to find optimal parameters $\theta$ for a generative sequential model such as a Transformer
\begin{equation}
X_i = \text{Transformer}(\mathcal{E}_i, \theta)
\label{eq:Transformer}
\end{equation}
that minimizes the loss function
\begin{equation}
\mathcal{L}(X, X^{\text{t}}) =  \sum_{i=1}^{|D|} - \mathcal{L}_{\text{pred}}(X_i, X^{\text{t}}_i) -\alpha f(X_i),
\label{eq:loss}
\end{equation}
where the first term is a measure of the model's ability to generate the expected ground-truth sequence $X^{\text{t}}$ and the second term corresponds to the design objective, weighted with $\alpha$. The model is trained using a dataset $D=\{X_i, X^{\text{t}}_i\}$.

Note that the generative model could not only be used to predict a solution to the problem on its own but also used in conjunction with traditional search methods to efficiently solve the problem. To this end, we propose hybrid methods by combining our Transformer-based model with two distinct search methods: Estimation of distribution algorithm (EDA) and Monte Carlo tree search (MCTS).

\subsection{Hybrid Methods}

\subsubsection{EDA+Transformer}

EDA is an evolutionary algorithm that uses a probabilistic model to iteratively sample and improve a population of solutions \cite{larranaga2012estimation}. We introduce a hybrid approach that combines EDA with a Transformer-based model to leverage the strengths of the two methods. In this method, EDA is used to explore the first few tokens of the sequence, which have the strongest influence on the whole sequence generation. Given the maximum sequence length of $N$, the method limits the sequence length considered by EDA to $N_{\text{EDA}} < N$ and then uses a Transformer-based model to complete the subsequent tokens before evaluating each candidate solution.

We use a bi-gram probabilistic model for EDA:
\begin{equation}
P(x_{1:N}) = \prod_{i=1}^N P(x_i|x_{i-1})
\end{equation}
where $x_1$ is the start token and subsequent tokens $x_i$ are generated using the conditional probability based on the previous token $x_{i-1}$.


\subsubsection{MCTS+Transformer}

MCTS is a heuristic search algorithm that balances the exploration of under-explored search paths with the exploitation of promising paths (\cite{coulom2006efficient, kocsis2006bandit}). Similar to the first method, we combine MCTS with a Transformer-based model to create a hybrid method. We limit the depth of the search tree explored by MCTS to $N_{\text{MCTS}} < N$, as MCTS is used to explore the critical initial tokens, while a Transformer-based model is used to complete the subsequent tokens before evaluating each candidate solution.

MCTS uses a heuristic function to decide which child node (token) to expand during the search, typically:
\begin{equation}
h = \frac{R_i}{v_i} + c \sqrt{\frac{\ln{V_i}}{v_i}}
\end{equation}
where $R_i$ is the accumulated reward for the $i$-th child, $v_i$ is the number of visits made at the child, $V_i$ is the total number of visits made at the parent, and $c$ is an exploration parameter (set to 1.4 in our case).


Both hybrid methods combine the strengths of search algorithms (EDA and MCTS), namely their exploration capabilities, with a Transformer-based model to generate high-quality solutions, potentially leading to more efficient exploration of the search space and better overall results.

\section{Domain-Specific Language} 
To represent valid gear train designs and help constrain the search space, a domain-specific language (DSL) is developed. A valid gear train sequence must conform to the grammar and lexicon of the DSL as follows. 

\subsection{Grammar} We define $\mathcal{S}=<$start$>$ and $\Sigma=<$end$>$. The set of variables enumerate the types of gear components and interfaces (the last two in the following list) considered, $\mathcal{V}=\{$ Shaft, Rack, Spur gear, Bevel gear, Miter gear, Worm gear, Hypoid gear, Translate, Mesh$\}$. The simplified set of rules $\mathcal{R}$ is shown in Table \ref{tab:grammar}, while the full grammar can be found in the supplemantary material. 

\subsection{Lexicon} The lexicon defines the possible tokens for each variable. For each component type, we assume that two to twelve different parts are available for an engineer (Table \ref{tab:catalgoue}), based on a gear component manufacturer's catalogue\footnote{\url{https://khkgears.net/new/gear_catalog.html}}. 

For the interface variable ``Translate'', we define a pair of translation tokens $tra+$ and $tra-$. They indicate toward which direction a shaft should be placed. The $+$ or $-$ sign indicates the direction relative to the current orientation. For example, if the current component is a spur gear oriented toward $[1, 0, 0]$, $tra+$ would represent the following shaft being placed toward $[1, 0, 0]$ while $tra-$ toward $[-1, 0, 0]$.

\begin{figure}[h!]
    \centering
    \begin{subfigure}[b]{0.23\textwidth}
        \centering
        \includegraphics[width=\textwidth]{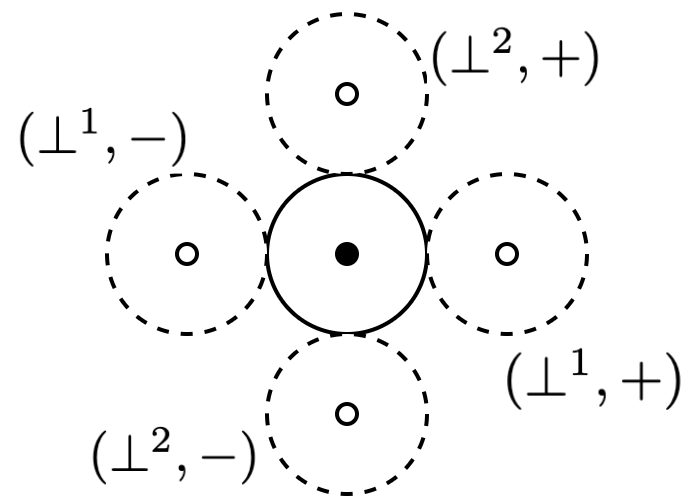}
        \caption{}
        \label{fig:image1}
    \end{subfigure}
    \hfill
    \begin{subfigure}[b]{0.234\textwidth}
        \centering

        \includegraphics[width=\textwidth]{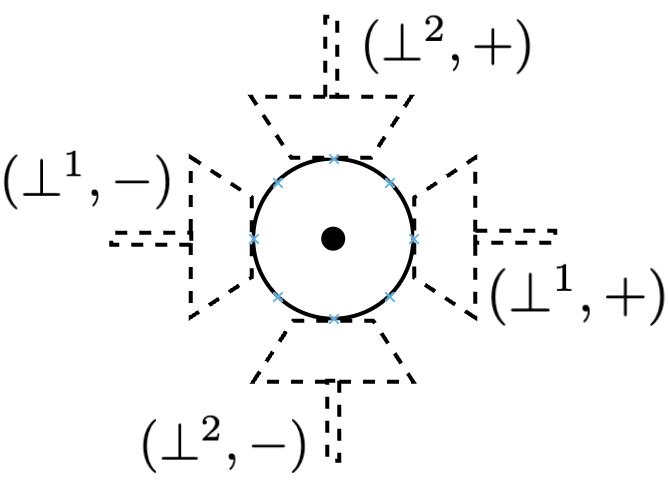}
        \caption{}
        \label{fig:image2}
    \end{subfigure}
    \hfill
    \begin{subfigure}[b]{0.24\textwidth}
        \centering
        \includegraphics[width=\textwidth]{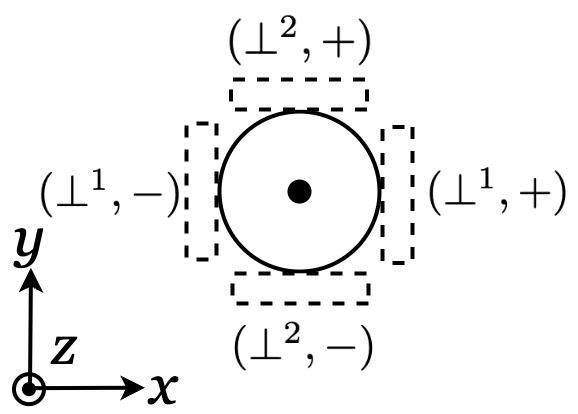}
        \caption{}
        \label{fig:image3}
    \end{subfigure}
    \hfill
    \begin{subfigure}[b]{0.2\textwidth}
        \centering
        \includegraphics[width=\textwidth]{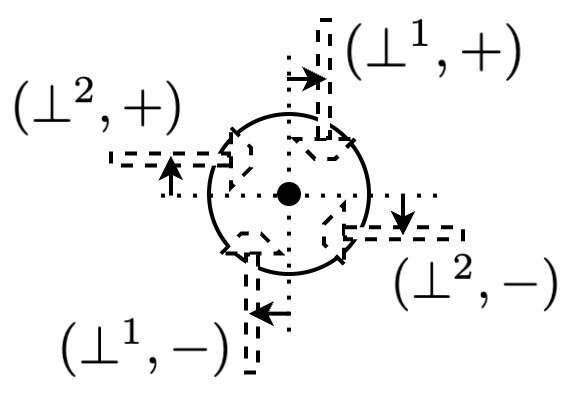}
        \caption{}
        \label{fig:image4}
    \end{subfigure}
    \caption{Examples of gear pair placements for each meshing token. (a) Spur gears. (b) Miter/Bevel gears. (c) Rack-and-pinion and worm gears.  (d) Hypoid gears.}
    \label{fig:four_images}
\end{figure}

Finally, we define a set of meshing tokens \{$(\perp^{1}, +)$, $(\perp^{1}, -)$, $(\perp^{2}, +)$, and $(\perp^{2}, -)$\} for ``Mesh'', which encode how the next gear component is placed relative to the current component (Figure \ref{fig:four_images}). The first element in the token tuple indicates along which axis the next component should be placed relative to the motion axis of the current component. Then, the second element indicates whether to place the component in the positive or negative direction. So for example, given a gear component with the motion axis of $[0, 0, 1]$, applying the token $(\perp^{1}, -)$ would result in the next component being placed along the direction of $[-1, 0, 0]$.

In total, the lexicon size is 52, consisting of 44 gear components, 6 interface tokens, and start and end tokens.

\begin{table}[h!]
\centering
\small
\begin{tabular}{p{1.3cm}p{6.2cm}}
\toprule
\textbf{LHS} & \textbf{RHS} \\
\midrule
 start & Translate-Shaft $\mid$ Rack-Mesh-Spur gear \\
Shaft & \parbox[t]{6.2cm} {\raggedright \emph{Gear}$^\dagger$-Mesh-\emph{Gear} $\mid$ Spur gear-Mesh-Rack $\mid$ end} \\
 Spur gear & Mesh-Spur gear $\mid$ Translate-Shaft $\mid$ end \\
 Rack & end \\
\emph{Gear} & Translate-Shaft $\mid$ end \\
\bottomrule
\end{tabular}
\caption{Simplified gear train grammar. LHS and RHS stand for left-hand side and right-hand side, respectively. $^\dagger$\emph{Gear} represents multiple component types listed in Table \ref{tab:catalgoue} excluding racks. Each variable type can be expanded with either part tokens for the component types or translation/meshing tokens for the interface types.}
\label{tab:grammar}
\end{table}

\begin{table}[h!]
\centering
\small
\begin{tabularx}{\columnwidth}{p{1.6cm}X}
\toprule
\textbf{Components} & \textbf{Part numbers / tokens} \\
\midrule
Shafts$^\dagger$ & SH-(*, 100, 200, 300, 400, 500) \\
Racks & \parbox[t]{5.9cm}{MRGF(1.5, 2, 2.5, 3)-500} \\
Spur gears & \parbox[t]{5.9cm}{MSGA1.5-(20, 40, 60, 80), MSGA2-(18, 25, 40, 60), MSGA2.5-(15, 40, 55, 70), MSGA3-(15, 30, 45, 60)} \\
Bevel gears & \parbox[t]{5.9cm}{SBSG2-(3020R, 2030L, 4020R, 2040L, 4515R, 1545L)} \\
Miter gears & MMSG2-20(R,L) \\
Worm gears & \parbox[t]{5.9cm}{\textit{Worm}: SWG1-R1 \\ \textit{Wheels}: AG1-(20R1, 40R1, 60R1)} \\
Hypoid gears & \parbox[t]{5.9cm}{\textit{Pinions}: MHP1-(3045L, 2060L, 1045L) \\ \textit{Rings}: MHP1-(0453R, 0602R, 0451R)} \\
\bottomrule
\end{tabularx}
\caption{Part numbers (used as tokens) considered for each component type obtained from the commercial catalogue. Abbreviations are used to combine similar part numbers with a consistent prefix; the numbers in parentheses indicate variations of the base part number. $^\dagger$Shafts are assumed to be pre-cut into different lengths at 100mm, 200mm, etc., denoted by the part number, or to place two gear components immediately side-by-side in the case of SH-*.}
\label{tab:catalgoue}
\end{table}

\section{Dataset Generation}
An important contribution of our work is a method to create the synthetic dataset required to train a Transformer model, named \textit{GearFormer} dataset, because there lacks any existing dataset of gear train designs. The dataset includes sequences that can be mapped to real-life gear trains paired with multiple requirement metrics evaluated with a physics simulator.
\subsubsection{Generate variable sequences}
Using the grammar in Table~\ref{tab:grammar} as production rules, we generated variable sequences having maximum 10 components to bound the problem but still produce realistic gear trains. A variable sequence with a maximum of 10 components can have up to 21 tokens, including the interface variables and the start/terminal tokens. An example variable sequence with five components is: $(<$start$>$, Rack (1), Mesh, Spur Gear (2), Translate, Shaft (3), Bevel Gear (4), Mesh, Bevel Gear (5), $<$end$>)$. In total, 37,606 unique variable sequences are generated.



\subsubsection{Generate token sequences}
For each variable in a variable sequence, one of the possible tokens is randomly chosen from our lexicon list. This generates a token sequence that resembles a real gear train design. For the variable sequence example above, a possible token sequence is $(<$start$>$, MRGF2-500, $(\perp^{2}, -)$, MSGA2-40, $tra-$, SH-200, SBSG2-3020R, $(\perp^{1}, +)$, SBSG2-2030L, $<$end$>)$. 

We discard sequences that are not physically feasible (i.e., at least two components interfere with each other) as explained in the following subsection. We generated a total of 7,363,640 grammatically valid and feasible sequences. Out of these, 0.05\% (3,681) were randomly selected for validation, another 0.05\% for testing, and the remaining sequences for training. Relatively small validation and test sets are used because each predicted sequence during validation/testing must be evaluated with a simulator to compute the actual requirement metrics, which can be time-consuming. 

\subsubsection{Physical feasibility check}
To verify the physical feasibility, we check for interference between all possible pairs of components in the sequence. We use the dimensions from the commercial catalogue to define a bounding box for each component, while keeping track of its current position and orientation. We check for intersections between the bounding box of the current component and those of the previous components, except for the one immediately preceding it, as they are supposed to be connected.

\subsubsection{Input encoding vector}

We construct a vector of the desired requirements with the following elements, referring to Eq. \eqref{eq:gear_opt}:
1) $\Tilde{\tau}_{\text{in}}$ = 1 for translation or 0 for rotation,
2) $\Tilde{\tau}_{\text{out}}$ = 1 for translation or 0 for rotation,
3) $\Tilde{s} \in \mathbb{R}$,
4-6) $\widetilde{\vec{p}} \in \mathbb{R}^3$,
7) the nonzero element index of $\widetilde{\vec{m}}$ as 0, 1, or 2, and
8) the sign of the nonzero element of $\widetilde{\vec{m}}$ as 1 or -1.


\subsubsection{Gear train simulator}

We implemented a simulator using Dymos\footnote{https://github.com/OpenMDAO/dymos} to compute all the requirement metrics given a gear train sequence. Dymos is an open-source Python library that enables simulation of time-dependent systems. The kinematics and spatial computations for each component type required were implemented as a Dymos \textit{component}. Given a sequence, a gear train system (defined as a \textit{group} in Dymos) is composed on-the-fly by instantiating the components corresponding to the parts in the sequence and connecting them based on the interface information. 




\section{GearFormer for Gear Train Synthesis}

\subsubsection{Model architecture}
GearFormer (Figure \ref{fig:model_arch}) is based on the standard Transformer architecture \cite{vaswani2017attention}. It includes a bidirectional encoder module that takes an input vector that encodes the design requirements, which are embedded with linear layers, ReLU activations, and batch normalization. The output of this encoder serves as context for the autoregressive decoder and is consumed via cross-attention. The decoder then predicts a sequence of output tokens, which represents a valid and feasible gear train design meeting the specified requirements while also minimizing the weight objective. Example designs generated with GearFormer are shown in Figure \ref{fig:designs}.


\subsubsection{Loss function}
Predicting each token of the sequence is a classification task that attempts to select the best class from a vocabulary of 53 tokens (52 tokens from the lexicon plus one for the end-of-sentence token), given the previous tokens and the context (design requirements). Hence, we use a standard cross-entropy loss for $\mathcal{L}_{\text{pred}}$ in Eq.~\ref{eq:loss}. We also include the weight of the gear train as the design objective loss and experiment with different ratios of $\alpha$.
\begin{equation}
\mathcal{L}(p(X), p(X^{\text{t}})) =  \sum_{i=1}^{|D|} - p(X^{\text{t}}_i) \log p(X_i) + \alpha f_w
\label{eq:loss_act}
\end{equation}

\subsubsection{Computing the weight objective with Gumbel-Softmax} 
Evaluating the weight objective in the training loop requires the ability to differentiably sample the output sequence from the model, and fetch the corresponding component weights from the catalogue.
To achieve this, we first extract the component weights from the gear catalogue, and store them in a coefficient vector $W_{\text{coef}}$.
Then, we use the straight-through Gumbel-Softmax~\cite{jang2016categorical} trick in the autoregressive sampling loop to generate each token in the sequence as a one-hot vector $s_i \in \{0, 1\}^{53}$.
By computing the dot product $s_i \cdot W_{\text{coef}}$, we can \textit{retrieve} the weight of the component represented by the token.
Repeating this for the entire output sequence and summing the weights gives us the value of the weight objective function.
This technique is quite general and can be used for any set of attributes associated to the tokens in the vocabulary.
Note that we calculate the weight of the shafts based on their length, assuming the material is carbon steel and the diameter of all the shafts is $0.01m$. We also assume that the weight of SH-* is $0kg$.

\subsubsection{Adaptive loss weighting} 
We set $\alpha = w_1 \cos(w_{\epsilon})$, where $w_1$ is a model parameter and $w_{\epsilon}$ is determined based on the epoch number $\epsilon$ as $w_{\epsilon} = \max(0, \frac{\pi}{2} - (\epsilon-1) \times \frac{\pi}{6})$
to bias the model on the cross entropy loss during the initial epochs. 

\section{Experiments}
First we present the metrics used to evaluate GearFormer plus its training and selection details. We then compare GearFormer and our hybrid methods against search methods on their own with a benchmark problem set.

\begin{table*}[ht!]
\small
\centering
\begin{tabular}{@{}l|lccccccccc@{}
  l 
  S[table-format=2.2(3)] 
  S[table-format=2.2(3)] 
  S[table-format=2.2(3)] 
  S[table-format=2.2(3)] 
  S[table-format=2.2(3)] 
  S[table-format=2.2(3)] 
  S[table-format=2.2(3)] 
  S[table-format=2.2(3)] 
 S[table-format=2.2(3)] 
 S[table-format=2.2(3)]
  @{}
  }
\toprule
&  \textbf{Model}& \textbf{Valid}$\uparrow$& \textbf{Feas}$\uparrow$& \textbf{Pos}$\downarrow$ & \textbf{Speed}$\downarrow$ & \textbf{Mot Vec}$\uparrow$ & \textbf{In-Mot}$\uparrow$ & \textbf{Out-Mot}$\uparrow$ & \textbf{Weight}$\downarrow$ & \textbf{Cand \#}$\downarrow$ \\
 & & \% & \% & m & log($\cdot$) & \% & \% & \% & kg & - \\
\midrule
\multirow{2}{*}{\rotatebox[origin=c]{90}{test}}
& GearFormer & 98.70 & \textbf{94.73} & \textbf{0.041} & \textbf{0.0229} & \textbf{99.04} & \textbf{100.0} & \textbf{99.92} & \textbf{6.25} & - \\
& Rand & - & 0.4766 & 0.860 & 1.9393 & 16.96 & 54.13 & 71.79 & 6.69 & - \\

\midrule
\multirow{5}{*}{\rotatebox[origin=c]{90}{30 random set}} 
& GearFormer & 96.67 & 93.33 & $\textbf{0.045}^{\pm0.047}$ & $0.0266^{\pm 0.042}$& \textbf{100.0} & \textbf{100.0} & \textbf{100.0} & $5.98^{\pm4.35}$ & \textbf{1}\\

& EDA$+$GF & - & \textbf{100.0} & $0.134^{\pm0.163}$ & $\textbf{0.0098}^{\pm0.016}$ & 83.33 & \textbf{100.0} & \textbf{100.0} & $3.26^{\pm3.59}$ & 10e3\\
& MCTS$+$GF & - & \textbf{100.0} & $0.209^{\pm0.362}$ & $0.0275^{\pm0.047}$ & 86.67 & \textbf{100.0} & \textbf{100.0} & $2.62^{\pm2.19}$ & 10e3\\
& EDA & - & \textbf{100.0} & $0.652^{\pm0.362}$ & $0.0930^{\pm0.132}$ & 80.00 & \textbf{100.0} & \textbf{100.0} & $\textbf{1.98}^{\pm1.55}$ & 10e4\\
& MCTS & - & \textbf{100.0}& $0.760^{\pm0.400}$ & $0.2175^{\pm0.308}$ & 66.67 &\textbf{100.0} & \textbf{100.0} & $3.20^{\pm3.09}$ & 10e4\\
\bottomrule
\end{tabular}
\caption{We compared GearFormer against multiple baselines. The upper section of the table compares GearFormer with the random baseline (Rand) on the test set, where the latter generates random sequences that conform to the grammar. The lower section of the table compares the metrics of GearFormer, hybrid methods, and search methods on their own, evaluated with 30 randomly sampled benchmark problems. `\textbf{Cand \#}' stands for the number of candidate solutions considered to find the best solution used to compute the metrics.}
\label{results12}
\end{table*}

\subsection{Evaluation Metrics}
We evaluated the model using several metrics based on Eq. \eqref{eq:gear_opt}. Validity and feasibility were assessed for the entire test set, while the remaining metrics applied only to valid sequences, as the simulator produces outputs solely for those.

\begin{itemize}[noitemsep, topsep=0pt, left=0pt]
    \item \textbf{Validity (Valid):} The percentage of sequences conforming to the grammar.
    \item \textbf{Feasibility (Feas):} The percentage of sequences that are valid and non-interfering.
    \item \textbf{Output position match (Pos):} Average Euclidean distance between target and actual positions $|\widetilde{\vec{p}} - \vec{p}(X)|$.
    \item \textbf{Speed ratio match (Speed):} Average RMSLE($\Tilde{s}, s(X)$), i.e., root mean squared logarithmic error to account for the wide range of speed ratios ($10^{-5}$--$10^{5}$).
    \item \textbf{Output motion vector match (Mot Vec):} The percentage of $\widetilde{\vec{m}} \cdot \vec{m}(X) = 1$.
    \item \textbf{Input/output motion type match (In-Mot/Out-Mot):} The percentages of $\Tilde{\tau}_{\text{in}} = \tau_{\text{in}}(X)$ and $\Tilde{\tau}_{\text{out}} = \tau_{\text{out}}(X)$.
    \item \textbf{Weight:} The average of $f_w$.
\end{itemize}

\subsection{GearFormer Model Training and Selection} We implemented our model using the x-transformers library\footnote{https://github.com/lucidrains/x-transformers} and PyTorch~\cite{Ansel_PyTorch_2_Faster_2024}. A Tesla V100-SXM2-16GB GPU was used for training and evaluation. Training was conducted for a maximum of 20 epochs, but stopped if the validation loss increased at any point. Each epoch requires over 200 seconds for evaluation with the simulator, resulting in a maximum of one hour per experiment. More details are provided in the supplementary materials.

\subsubsection{Loss function} 
\begin{figure}[!h]
    \centering    \includegraphics[width=0.9\linewidth]{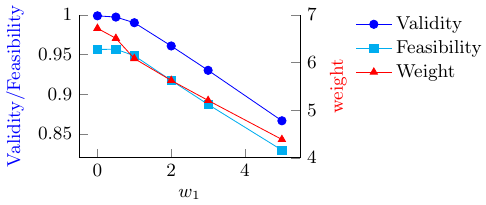}
    \caption{As we increase $w_1$ to prioritize the weight loss term, the average weight of the generated sequences drops at the expense of producing fewer valid and feasible sequences.}
    \label{fig:weight_loss}
\end{figure}

We experimented with various values of ($w_1$) for our objective function weighting. As shown in Figure ~\ref{fig:weight_loss}, as $w_1$ increases, the average weight of the generated sequences drops; however, the model generates fewer valid and feasible sequences. Note that the average weight for the ground truth sequences in the validation set is $6.72kg$. We chose $w_1=1.0$ for our model to prioritize generating valid and feasible sequences for our experiments.

\subsubsection{Model size} We experimented with different values for dimension, depth, and attention heads on the validation set (see the supplementary materials). To balance having fewer parameters with achieving good results, we decided on the dimension of 512, depth of 6, and 8 attention heads.

\subsection{GearFormer Against and With Search Methods}

We evaluate GearFormer and the two hybrid methods developed, named EDA+GF and MCTS+GF, versus the search methods on their own. Because it is impractical to run search methods on the entire test dataset, we randomly sampled 30 benchmark problems from the test set. We set both $N_{\text{EDA}} = 6$ and $N_{\text{MCTS}} = 6$.

\subsection{Results}
The upper section of Table~\ref{results12} compares the performance of GearFormer with the random baseline (Rand) method on the test set. The baseline method randomly generates sequences that conform to the grammar. GearFormer significantly outperforms the baseline across all metrics. The results for the baseline also highlight the difficulty of finding feasible solutions that address multiple design requirements. 

The lower section of Table~\ref{results12} shows that GearFormer outperforms or is at par with EDA and MCTS in four out of the five design requirements. Because the search methods are specifically designed to only generate valid sequences and can pick only the feasible solutions out of multiple candidates, they achieve 100\% accuracy on those metrics. They can also find more lightweight solutions, at the expense of not addressing the design requirements. It should be emphasized that GearFormer can instantly generate a high-quality solution, compared to the search methods that selected the best solution out of 10,000 candidate solutions evaluated. 

Table~\ref{results12} also highlights the benefit of our hybrid methods. Compared to the search methods on their own, all design requirement metrics improved with the hybrid methods (in fact, the speed metric is better than GearFormer's). This came at the expense of the weight objective for EDA+GF but for MCTS+GF, the weight objective even improved slightly. Lastly, the hybrid methods only used 1/10$^\text{th}$ of the candidate solutions compared to the search methods on their own to achieve such results, demonstrating the improved efficiency.

Note that the average inference time for GearFormer to generate a single candidate solution were 0.328s. For the search methods, the average evaluation time for a single candidate solution was 0.039s (which includes simulation), or about 6.5 minutes to perform a single run of EDA or MCTS.


\section{Discussion and Conclusions}

Our work has showcased the possibility of solving a challenging mechanical configuration design problem, particularly gear train synthesis, with a Transformer-based model named GearFormer. GearFormer not only generates solutions instantly but also outperforms traditional search methods in finding quality solutions that meet design requirements. This is tremendously valuable in assisting engineers to quickly explore multiple solutions within a shorter amount of time. Additionally, GearFormer could be used to auto-complete a partial gear train design provided by the engineer,  either by suggesting or ranking potential lists of subsequent components in real-time. This enables an interactive workflow between the engineer and the tool that is unattainable with traditional search methods.

We also introduce hybrid search methods by combining EDA and MCTS with GearFormer. These methods leverage the explorative capability of search methods for early decisions during sequence generation and the generative model's ability to complete the remaining decisions toward high-quality solutions. The hybrid methods found solutions that better address design requirements than the search methods on their own within a shorter amount of time.

Another important contribution of our work is the introduction of the first dataset on gear train synthesis. We developed a domain-specific language (DSL) and a physics simulator that can be used to generate more datasets in the future. Our language can be extended to incorporate additional grammar and lexicon to produce richer datasets that can be used to train a model for more complex gear train designs.

Finally, this work elucidates how a difficult engineering design problem can be formulated and solved using a deep generative model approach and provides a strong evidence that the approach can be effectively utilized in engineering design. The modular nature of our methodology allows for extension to a wide range of configuration design problems, such as hydraulic systems, modular robots, frame structures, etc., by leveraging appropriate DSLs and physics simulators for the respective application. We believe that many research opportunities exist in this space with notable implications for societal innovation.

\bibliography{aaai25}

\end{document}